\documentclass[letterpaper, 10 pt, conference]{ieeeconf}  
\IEEEoverridecommandlockouts                              
\usepackage{graphics} 
\usepackage{epsfig} 
\usepackage{subfigure}
\usepackage{mathptmx} 
\usepackage{times} 
\usepackage{amsmath} 
\usepackage{amssymb}  
\usepackage{algorithm2e}
\usepackage{multirow}

\usepackage{cite}
\usepackage[
colorlinks=true,
linkcolor=black,
urlcolor=blue,
citecolor=black
]{hyperref}

\title{\LARGE \bf
Learn the Manipulation of Deformable Objects Using Tangent Space Point Set Registration 
}

\author{Rui Wang$^{1}$, Te Tang$^{2}$ and Masayoshi Tomizuka$^{2}$
\thanks{$^{1}$
Department of Mechanical Engineering, Tsinghua University, Beijing, China.
        {\tt\small wangrui15@mails.tsinghua.edu.cn}
}%
\thanks{$^{2}$
Department of Mechanical Engineering, University of California, Berkeley, CA, USA.
        {\tt\small \{tetang, tomizuka\}@berkeley.edu}
}%
\thanks{This work is supported in part by NSF(Award \#1734109). Rui Wang received a scholarship from Tsinghua University.
}%
}
\begin{document}

\maketitle
\thispagestyle{empty}
\pagestyle{empty}

\begin{abstract}
Point set registration is a powerful method that enables robots to manipulate deformable objects. By mapping the point cloud of the current object to the pre-trained point cloud, a transformation function can be constructed. The manipulator's trajectory for pre-trained shapes can be warped with this transformation function, yielding a feasible trajectory for the new shape. However, usually this transformation function regards objects as discrete points, and dismisses the topological structures. Therefore, it risks over-stretching or over-compression during manipulation. To tackle this problem, this paper proposes a tangent space point set registration method. A tangent space representation of an object is constructed by defining an angle for each node on the object. Point set registration algorithm runs in this newly-constructed tangent space, yielding a tangent space trajectory. The trajectory is then converted back to Cartesian space and carried out by the robot. Compared to its counterpart in Cartesian space, tangent space point set registration is safer and more robust, succeeding in a series of experiments such as rope straightening, rope knotting, cloth folding and unfolding.

\end{abstract}

\section{Introduction}
Robotic learning from human demonstration is a powerful method for robotic arms to acquire a certain set of skills. Based on the idea that intelligent beings such as human acquire new skills by ``imitation'', robots are programmed to imitate their trainers in a certain task \cite{learn_from_demo}. For example, after one round of training by human, robots can learn to tie knots with different ropes and fold clothes of various sizes.

In previous works \cite{tang2018ral, tang2018ijrr, Schulman2016, hclin2018grasping}, it is found that point set registration  algorithms are promising in robotic learning from human demonstration. Point set registration is a process to construct a transformation function that maps one point set to another. Regarding its application in robotic learning, robots are first led by humans to manipulate a baseline object. The trajectory information during this training process is recorded by the robot. When faced with a similar object, the robot is able to do a quick mapping of the previous object to the current one by point set registration algorithm, obtaining a transformation function, and warp the recorded training trajectory with this function. The warped trajectory will be carried out to manipulate the new object. In this paper, the process of human demonstration is called ``training'', and the process of manipulating a new object is called ``testing''. \par
\begin{figure}[t]
\centering
\includegraphics[scale=0.55]{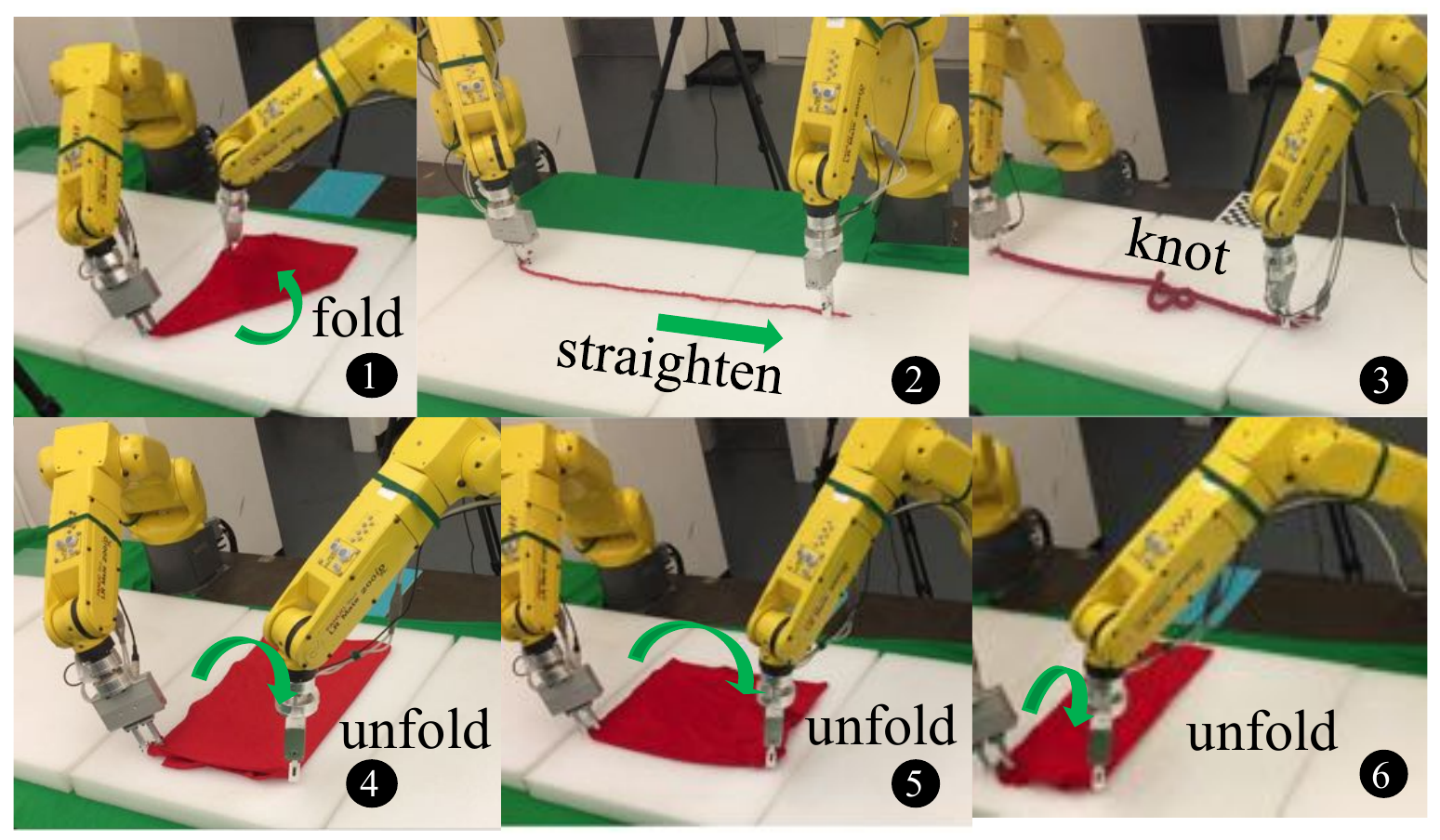}
\caption{
The dual-arm robot system carries out a series of tasks. 1-3: cloth folding, rope straightening, rope knotting. 4-6: the robot was trained to unfold a piece of cloth, and transfers the skill to unfold normal, wider, and narrower cloth, respectively.
}
\label{fig_expRobot}
\end{figure}
While point set registration most often finds its application in data alignment and information recovering, in this work it is employed to recognize object features and assign their correspondence accordingly. The method \cite{Schulman2016} has achieved success in many different experimental scenarios.\par
One unsolved problem is that the robot considers data collected by the camera as discrete points, and the warping function loses the object's length feature. Take rope manipulation as an example. A rope is modeled as a point set which is uniformly distributed along the rope. Physics laws require that the distance between any adjacent points on a rope remains unchanged throughout the manipulation. However, a registered point set does not necessarily satisfy this condition. Therefore in some cases, the robot might over-stretch the object, jeopardizing the manipulation. In previous works, we proposed a new algorithm\cite{TSM-RPM} to tackle this problem. A ``tangent space'' was constructed and point set registration was run in tangent space, after which the tangent information was integrated along the length of the rope, yielding its coordinates in Cartesian space. Simulation results showed that the algorithm successfully avoided over-stretching and over-compression. In this work, a modified method is proposed using the same basic idea. The algorithm is verified in experimental settings with various applications. Experiments prove that our algorithm is able to generate a reasonable trajectory and carry out a manipulation whose result desirably resembles the training scene, while avoiding over-stretching and over-compression. The algorithm works successfully both on ropes and cloths. This paper presents the modified algorithm, offering detailed implementation notes, analysis, and experimental validation. \par

The remainder of this paper is organized as follows: Section II introduces related works of point set registration, as well as their novel application potential in robotic learning from human demonstration. Section III focuses on the coherent point drift (CPD) algorithm, and its application in robotic manipulation without considering the physical constraints of a soft body. Section IV elaborates on the proposed new algorithm, tangent space point set registration. Section V offers an insight into the experimental platform, with details of its system architecture. Section VI introduces experimental results and analysis.

Supplementary videos can be found at \cite{ExperimentVideo}.

\section{Related Works}
Point set registration is an algorithm developed to find a transformation function that registers one point set to another. Some of the well-known algorithms include iterative closest point (ICP\cite{ICP}), thin plate spline-robust point matching (TPS-RPM\cite{TPS-RPM1}) and coherent point drift (CPD\cite{CPD}). The ICP algorithm assigns a binary correspondence between points and finds a rigid transformation function. TPS-RPM and CPD, on the other hand, apply soft correspondence with a probabilistic view, and can find non-rigid functions that register points better.

Though point set registration is not designed directly for robotic learning from demonstration, it has proven to be a powerful tool. In \cite{Schulman2016}, the TPS-RPM algorithm was utilized to teach robots to manipulate ropes. The point cloud of the training rope is mapped to the point cloud of the testing rope through TPS-RPM, and the same function is used to warp the training trajectory, hence obtaining the trajectory in test. \cite{lee2014unifying} and \cite{lee2015learning} extended this work by considering the constraints imposed by robot's joints, obstacles in the space, and object's surface normal. In \cite{tang2017state}, the CPD algorithm is proposed to track deformable object in real time by registering the state estimation to the measured point cloud. A uniform framework for state estimation, task planning and trajectory planning for deformable object manipulation was also presented in \cite{tang2018ral} by utilizing CPD registration. 

It should be noted that most of the previous methods did not consider the geometric constraints of the manipulated object. Point set registration regards a physical object as a collection of discrete points and may therefore extend or shrink the object. We formerly proposed a method, TSM-RPM\cite{TSM-RPM} which tackles this problem in rope manipulation. TSM-RPM fits a thin plate spline in tangent space, and integrates the sinusoidal and cosinoidal values of the rope angle along the rope length to get the Cartesian coordinates. TSM-RPM proves effective in rope straightening and wire harnessing in simulation, but is not yet verified with experiments. In this work, the general point set registration algorithm is applied in tangent space to conserve the object length. As a specific implementation, the point set registration process makes use of coherent point drift algorithm\cite{CPD} which, compared with previous algorithms, can be extended to higher dimensions and in general has a better registration performance under occlusion. Our work also for the first time realizes this idea in experiments, both with ropes and cloths.
  
\section{Manipulation by Coherent Point Drift}
\subsection{Coherent Point Drift Method}

The coherent point drift(CPD) method was proposed to perform point set registration between similar point sets. The basic idea is to apply a probabilistic view in registration, and find out a ``soft'' instead of binary correspondence between points. Given two point sets, we divide them into \textit{observation set} and \textit{reference set}. The reference point set is transformed by function $f$, resulting in a \textit{transformed reference set}. By maximizing the probability of the observation point set data being collected from the transformed reference point set, an optimal $f$ is found.

Given two point sets, the observation group $\mathbf{X}$ and the reference group $\mathbf{Y}$, the goal is to register $\mathbf{Y}$ to $\mathbf{X}$ with function $f$. $\mathbf{X} = \{ \mathbf{x}_1, \mathbf{x}_2, \dots , \mathbf{x}_N \} \in \mathbb{R}^{N \times D}$, where $\mathbf{x}_n \in \mathbb{R}^D$ is the position of the $n$th node. $\mathbf{Y} = \{\mathbf{y}_1, \mathbf{y}_2, \dots , \mathbf{y}_M \} \in \mathbb{R}^{M \times D}$.  Let $f(\mathbf{Y}) = \mathbf{Z}$. To measure the precision of the transformation, calculate the probability that $\mathbf{X}$ is an observation of $\mathbf{Z}$ with a Gaussian mixture model(GMM). Note that the same weight is assigned for each point in $\mathbf{Z}$ in GMM. For a GMM of uniform variance $\sigma$, we have 
\begin{align}
  p(\mathbf{x}_n) = \frac{1}{M} \sum_{m = 1}^{M} \frac{1}{ (2\pi\sigma^2)^{\frac{D}{2}} } exp(-\frac{{\Vert \mathbf{x}_n - \mathbf{z}_m \Vert}^2}{2\sigma^2})
  \label{eq:G}
\end{align}

To further deal with the problem of noise, an additional uniform distribution is introduced, with a uniform probability density of $\frac{1}{H\times W}$, where $H$ and $W$ stand for the total height and width of the area that is being observed (e.g. by the camera). Let $A = H \times W$. With this modification, we obtain 
\begin{align}
  p(\mathbf{x}_n) = \omega \frac{1}{A} + (1 - \omega) \frac{1}{M} \sum_{m = 1}^{M} \frac{1}{ (2\pi\sigma^2)^{\frac{D}{2}} } exp(-\frac{{\Vert \mathbf{x}_n - \mathbf{z}_m \Vert}^2}{2\sigma^2})
  \label{eq:uniform}
\end{align}
where $\omega$ is some coefficient that is set manually, indicating the level of noise.

According to the maximum likelihood estimation, the goal is to maximize the total likelihood of $\mathbf{X}$ being the observation result of $\mathbf{Z}$. The total likelihood is defined as the multiplication of $p(\mathbf{x}_n)$ for each individual $\mathbf{x}_n$ in $\mathbf{X}$. Let $N$ denote the size of $\mathbf{X}$. Take the negative natural logarithm of this likelihood multiplication, and the expression that needs to be minimized is obtained:
\begin{align}
  L = -\sum_{n = 1}^{N} ln\big[(1- \omega)\sum_{m = 1}^{M}\frac{1}{M}p(\mathbf{x}_n|\mathbf{z}_m) + \omega\frac{1}{A}\big]
  \label{eq:Qln}
\end{align} where 
\begin{align}
  p(\mathbf{x}_n|\mathbf{z}_m) = \frac{1}{ (2\pi\sigma^2)^{\frac{D}{2}} } exp(-\frac{{\Vert \mathbf{x}_n - \mathbf{z}_m \Vert}^2}{2\sigma^2})
  \label{eq:p(x|y)}
\end{align}

In the application of CPD, a regularization term is required that prevents the function $f$ from getting too arbitrary. This is important as the desired transformation should register adjacent points to roughly adjacent positions. This \textit{coherence} is measured by the Fourier domain norm\cite{Fourier} of the function. Suppose $f(x)$ is a transformation where $f(x) = x + v(x)$. Its Fourier-domain norm is $\int_{\mathbb{R}^D} \frac{|V(s)|^2}{G(s)} ds$, where $V(s)$ is the Fourier transform of $v$ and $G(s)$ is a symmetric filter with $G(s) \rightarrow 0$ as $s \rightarrow \infty$. The larger the Fourier-domain norm, the less smooth the function $v$. Detailed mathematical explanation can be found in \cite{Fourier}\cite{Fourier2}. Applying this regularization, the expression of cost function $Q$ that needs to be minimized is obtained
\begin{align}
  Q(\sigma^2, f) = -\sum_{n = 1}^{N} ln\big[(1- \omega)\sum_{m = 1}^{M}\frac{1}{M}p(\mathbf{x}_n|\mathbf{z}_m) + \omega\frac{1}{A}\big] + \frac{\lambda}{2}\phi(f)
  \label{eq:Qregularized}
\end{align}
where 
\begin{align}
  \phi(f) = \int_{\mathbb{R}^D} \frac{|V(s)|^2}{G(s)} ds
  \label{eq:phi(f)}
\end{align}
In our specific implementation, in \eqref{eq:phi(f)}, $G(s)$ is chosen as a Gaussian.

Also for $\phi(f)$, define an Operator $\Phi$ such that 
\begin{align}
  \phi(f) = \Vert \Phi v \Vert ^2
  \label{eq:PhiOperator}
\end{align}
This operator will be of use in future derivation steps.

Note that $\sigma^2$ and $f$ are both unknown and need to be determined, while $\lambda$ and $\omega$ are manually assigned.

Next, apply the classic EM algorithm\cite{EM} to find $f$ and $\sigma^2$.
\vspace{\baselineskip}

\textbf{E-step $i$}\\
The E-step uses previously achieved result to calculate $p^{i-1}(m|\mathbf{x}_n)$, which is a posterior probability describing the chance that $\mathbf{x}_n$ is an observation of $\mathbf{z}_m$ or noise(when $m = M+1$). Using Bayes' theorem and the law of total probability, this probability can be calculated (the $i-1$ superscript will be left out for convenience):

\begin{align}
	p(m|\mathbf{x}_n)
    = \frac{p(\mathbf{x}_n|m)p(m)}
      {\sum_{i = 1}^{M+1}p(\mathbf{x}_n|i)p(i)}
    \label{eq:bayes}
\end{align}
where $p(\mathbf{x}_n|m)$ and $p(m)$ are as defined in \eqref{eq:p(m)} and \eqref{eq:p(x|m)} except that all probabilities are in step $i-1$ instead of step $i$.

\vspace{\baselineskip}

\textbf{M-step $i$}\\
The M-step tries to minimize Q with respect to $\sigma ^2$ and $f$. However, for the original expression \eqref{eq:Qln} where $L$ is in the form of $ -\sum_n ln(\sum_m p_{n,m}) $, no closed-form maximization solution is available. In order to deal with this problem, apply Jensen's inequality\cite{jensen} such that an upper bound of $L$ is found as
\begin{align}
  \tilde{L} = -\sum_{n = 1}^{N} \sum_{m = 1}^{M + 1}p^{i-1}(m|\mathbf{x}_n)ln\big[p(m)p^i(\mathbf{x}_n|m)\big]
  \label{eq:Qupperbound}
\end{align} where
\begin{equation}
p(m) = \left\{
\begin{array}{cc}
	(1-\omega)\frac{1}{M},  & m = 1,\dots, M  \\
	\omega, & m = M+1  \\
\end{array}
\right.
  \label{eq:p(m)}
\end{equation}
\begin{equation}
p^i({\mathbf{x}_n} | m) = \left\{
\begin{array}{cc}
	\frac{1}{ (2\pi\sigma^2)^{\frac{D}{2}} } exp(-\frac{{\Vert \mathbf{x}_n - \mathbf{z}_m^i \Vert}^2}{2\sigma^2}) ,  & m = 1,\dots, M  \\
	\frac{1}{A}, & m = M+1  \\
\end{array}
\right.
  \label{eq:p(x|m)}
\end{equation}

Instead of minimizing $L$, which is mathematically difficult, we minimize $\tilde{L}$. Further adding the regularization term, the final version of the function that calls for optimization is obtained
\begin{align}
  \tilde{Q}(\sigma^2, f) = -\sum_{n = 1}^{N} \sum_{m = 1}^{M + 1}p^{i-1}(m|{\mathbf{x}_n})ln\big[p(m)p^i(\mathbf{x}_n|m)\big] + \frac{\lambda}{2}\phi(f)
  \label{eq:Q'final}
\end{align}

Substitute \eqref{eq:p(x|y)} into \eqref{eq:Q'final} and ignore constants independent of $f$ and $\sigma$, and equation \eqref{eq:Q'final} can be simplified as
\begin{equation}
\begin{aligned}
\tilde{Q}^{i-1} = &\frac{1}{2\sigma^2}\sum_{n = 1}^M p^{i-1}(m|\mathbf{x}_n)\Vert \mathbf{x}_n - f^{i-1}(\mathbf{y}_m) \Vert ^2 \\
&+ \frac{D}{2}\sum_{n = 1}^N\sum_{m = 1}^M p^{i-1}(m|\mathbf{x}_n)ln\sigma^2  + \frac{\lambda}{2}\phi(f)
\end{aligned}  
\end{equation}
where $\tilde{Q}^{i-1}$ is a function of $(\sigma^2)^{i-1}$ and $f^{i-1}$.

It can be proved by calculus of variations that the function $f$ which minimizes $\tilde{Q}$ is in the following form:
\begin{align}
  f(\mathbf{x}) = \mathbf{x} + \sum_{m = 1}^{M}\mathbf{w}_mG(\mathbf{x}, \mathbf{\mathbf{y}_m})
  \label{eq:Green's function}
\end{align}
where $\mathbf{w_m} = \frac{1}{\sigma^2\lambda}\sum_{n = 1}^Np^{i-1}(m|\mathbf{x}_n)(\mathbf{z}_m - f(\mathbf{y}_m))$. Recall the operator $\Phi$ defined in \eqref{eq:PhiOperator}. $G$ is a Green's function of $\hat{\Phi}\Phi$, with $\hat{\Phi}$ being the adjoint operator to $\Phi$.

Also define $\mathbf{W} = [\mathbf{w}_1,...,\mathbf{w}_M]^T$ so that everything can be written in matrix form. Now with this form, rewrite $\tilde{Q}$ as a function of $\mathbf{W}$ and $\sigma^2$, whereafter both $\mathbf{W}$ and $\sigma^2$ can be updated by setting $\tilde{Q}$'s partial derivatives with respect to either of them to zero. The mathematical details can be found in\cite{CPD}.

Finally, function $f$ is updated as
\begin{align}
  f^i(\mathbf{Y}) = \mathbf{Y} + \mathbf{GW}
\end{align}
where $\mathbf{G}$ is a kernel matrix with $\mathbf{G}_{i, j} = exp(-\frac{1}{2}\Vert \mathbf{y}_i - \mathbf{y}_j \Vert ^2)$.

\subsection{Application in Cartesian Space Trajectory Transformation}
As CPD algorithm can register one point set to another with a function $f$, it was proposed to do scenario warping in Cartesian space. In this section, this idea will be revisited.

Consider that in training and testing, two point sets of objects, $\mathbf{Y}$ and $\mathbf{X}$ are given, respectively. Now use the algorithm described above to obtain a transformation function, $f$, which warps $\mathbf{Y}$, such that $\mathbf{X}$ is close to the warped $\mathbf{Y}$.

Note that during the training, the trajectory of the robot end-effector $\mathbf{T}$, was also recorded. $\mathbf{T}$ is a vector storing the coordinates of robot's end effector at each critical time frame. Transforming $\mathbf{T}$ with $f$ yields $\mathbf{T'} = f(\mathbf{T})$. $\mathbf{T'}$ is taken as the new trajectory for robotic manipulation at test.

\begin{figure}
\centering
\includegraphics[scale=0.4]{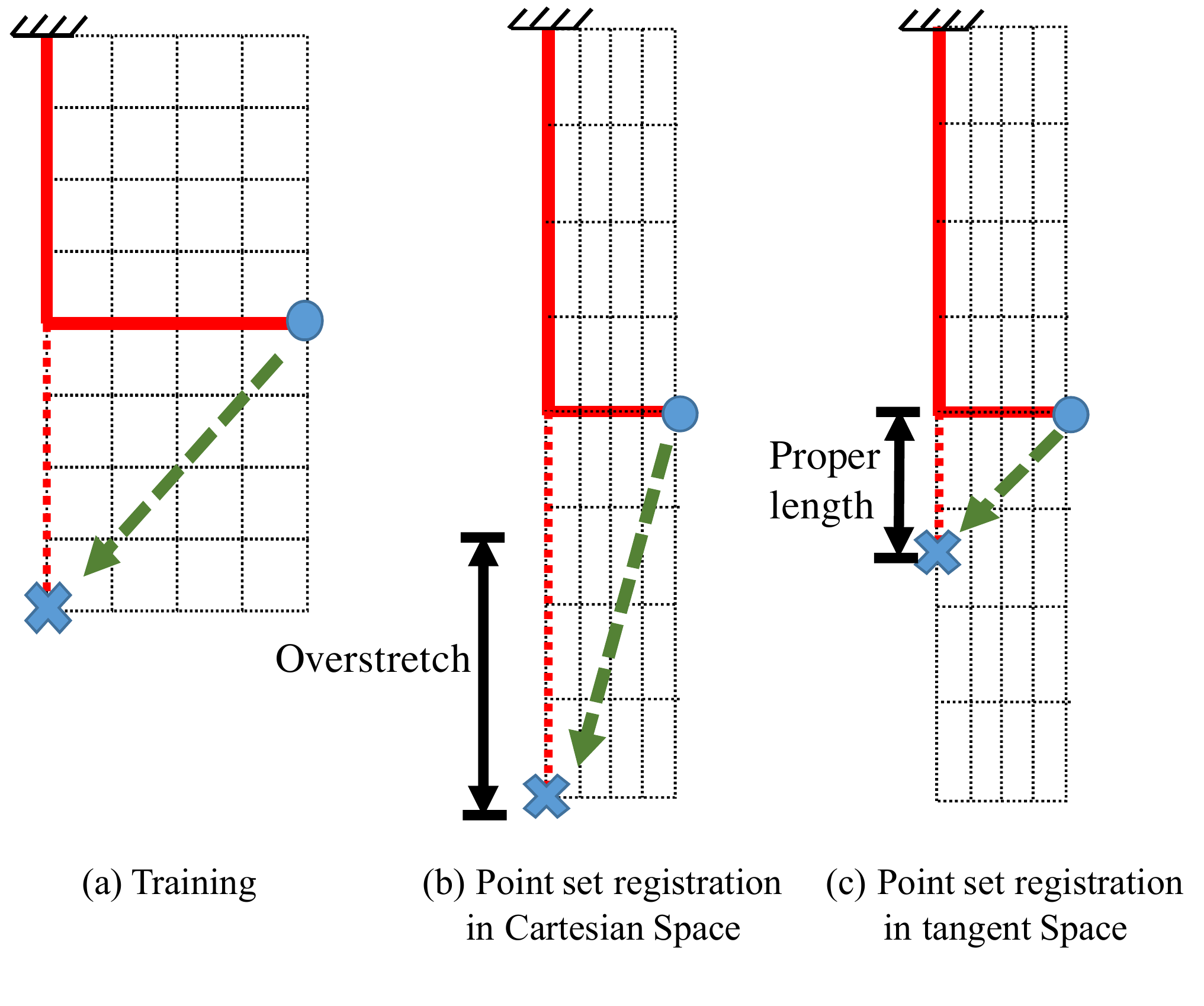}
\caption{Comparison of rope straightening task by methods in Cartesian space and in tangent space \cite{TSM-RPM}. Solid line shows the deformable rope, and dashed line is the manipulation trajectory. One end of the rope is fixed at the origin. The point set registration algorithm in Cartesian Space tries to register the two scenarios by shrinking Cartesian space in the horizontal direction while expanding in the vertical direction. As a result, it shrinks/expands the training trajectory in the same way to get the test trajectory (green line in (b)). This test trajectory, however, violates physical constraints and over-stretches the rope during manipulation. Point set registration in tangent space, on the other hand, avoids this problem, as shown in (c).}
\label{fig_comparison1}
\end{figure}

This algorithm works well on tasks like rope knotting. However, it has serious problems when it comes to rope straightening. This is because point set registration methods see an object as a set of discrete points. Transformation $f$ does not preserve the distances between adjacent points, even though in practice this conservation is required. An example is illustrated in Fig.~\ref{fig_comparison1}(b). Not meeting this physical requirement can result in failure or even damage of the object if it is to be stretched to its full length in manipulation.

\section{Trajectory Generation Algorithm}
\subsection{Trajectory Generation in Tangent Space}
In order to solve the problem of over-stretching/over-compression, a new method of trajectory generation is proposed, which applies point set registration in tangent space.
 
\begin{figure}[h]
\centering
\includegraphics[scale=0.45]{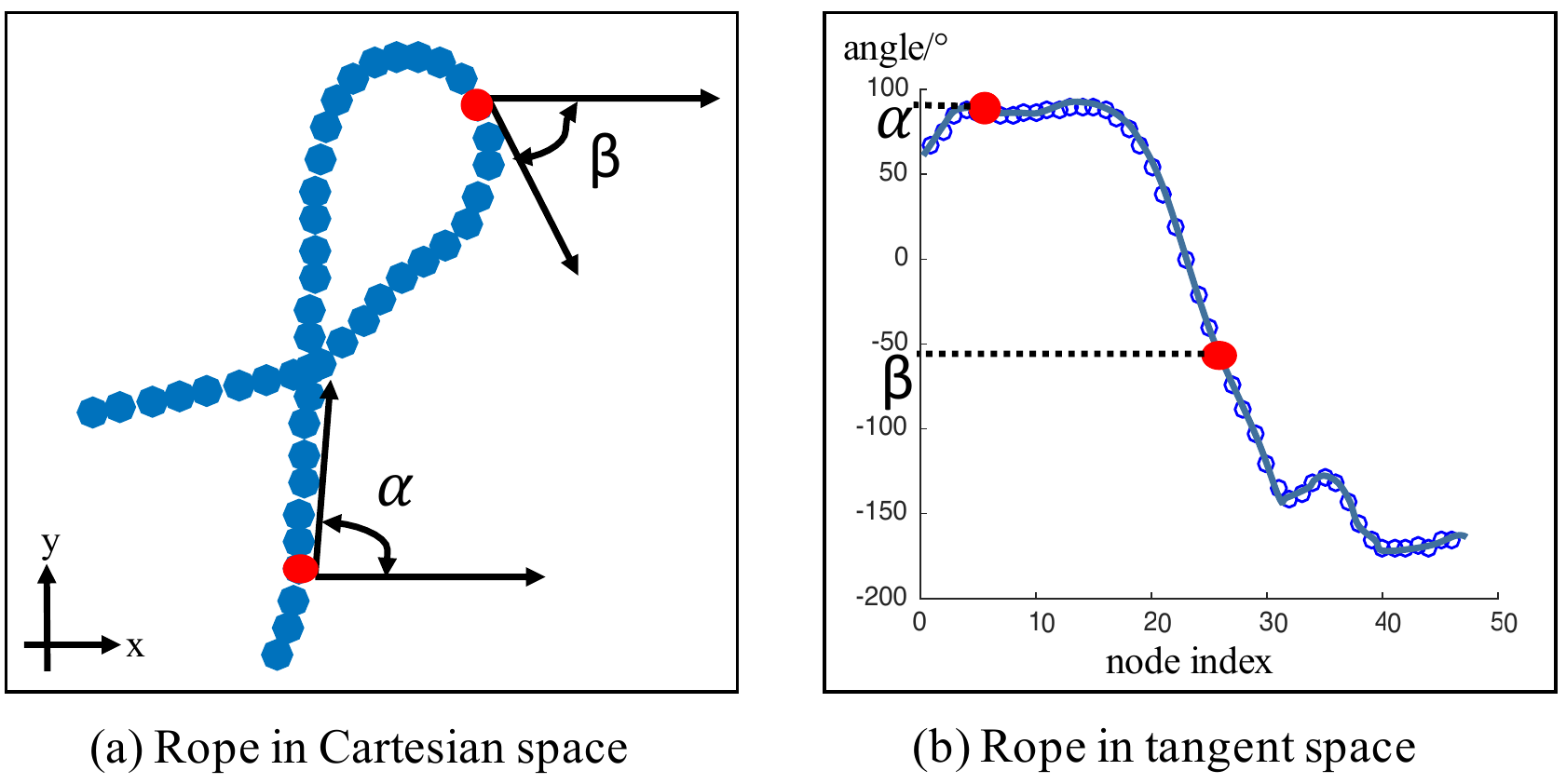}
\caption{A rope in (a) is converted into its tangent space representation in (b). Red nodes are examples which show how the angles are calculated.}
\label{fig:cartesian2tangent}
\end{figure}

First, the tangent information of a rope needs to be extracted. Let $\theta$ denote the angle between a node on the rope and the horizontal positive direction. A curve of $\theta$ as a function of node index can therefore be plotted. An example of this Cartesian-tangent conversion is shown in Fig.~\ref{fig:cartesian2tangent}. Note that the scale of $x$ axis can be modified arbitrarily without losing any essential information.
 Let us regard this plot in Fig.~\ref{fig:cartesian2tangent} as the tangent space representation of the rope. In a further step, combine all the $\theta$ into a vector and name it $\mathbf{X}_{\theta}$ for training set and $\mathbf{Y}_{\theta}$ for test set.

Next, instead of applying point set registration in Cartesian space, calculate the warping function $f$ in tangent space. Performing $f$ on the new vector $\mathbf{Y}_{\theta}$ yields $\mathbf{Z}_{\theta}$. Now calculate a correspondence matrix between $\mathbf{Z}_{\theta}$ and $\mathbf{X}_{\theta}$, so that the correspondence between each point pair in $\mathbf{X}_{\theta}$ and $\mathbf{Y}_{\theta}$ can be obtained. As $\mathbf{X}$ is considered as an observation of $\mathbf{Z}$, the probability that $\mathbf{x}_{\theta n}$ is an observation of each $\mathbf{z}_{\theta m}$ can be calculated. This is a posterior probability, with the condition that $\mathbf{x}_{\theta n}$ is observed. With CPD algorithm, in the E-M steps, this probability has already been derived \eqref{eq:bayes}.

\begin{figure}[h]
\centering
\includegraphics[scale=0.55]{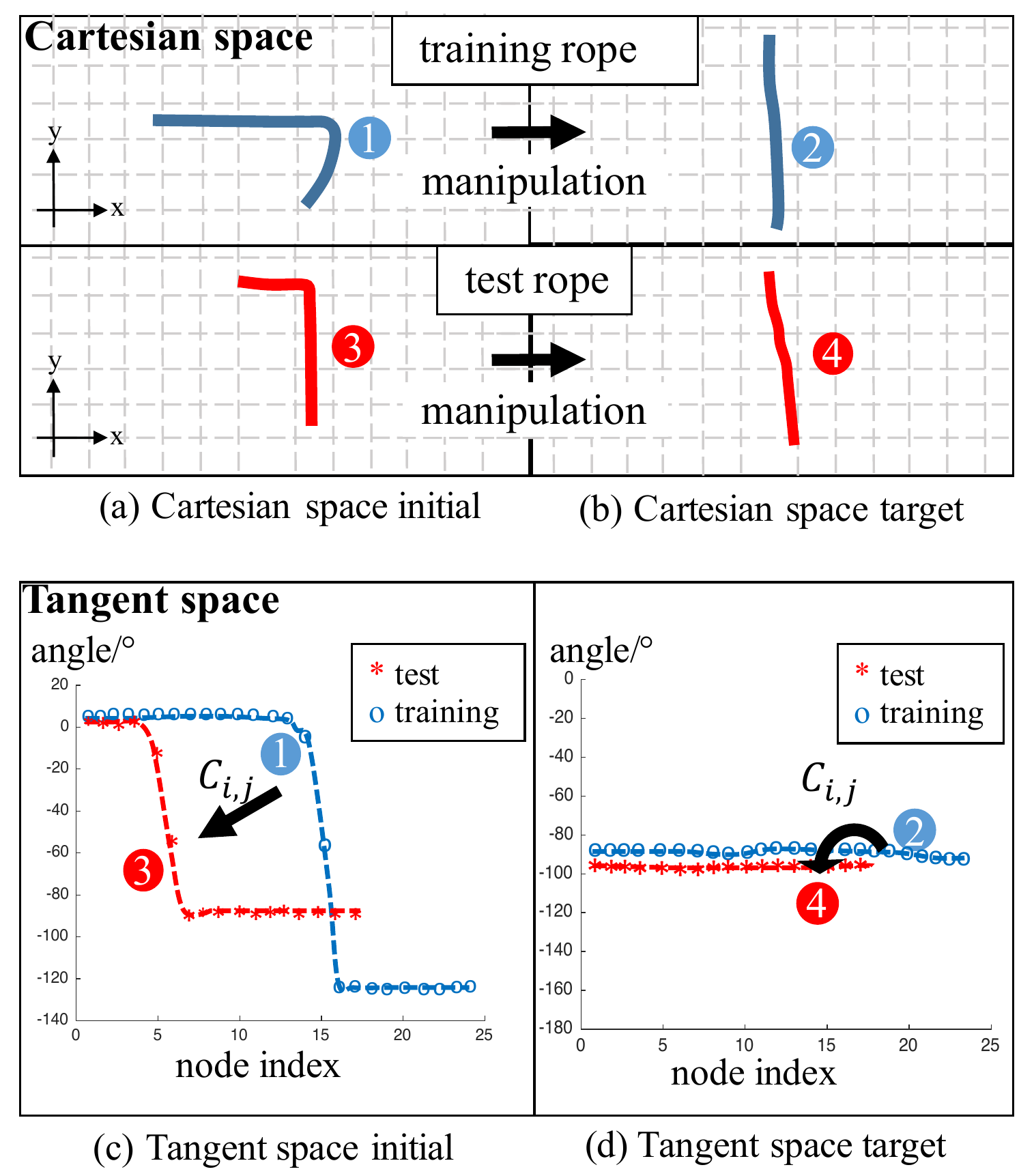}
\caption{A demonstration of tangent space point registration on rope straightening task. (a)(b), (c)(d) are Cartesian space representation and tangent space representation of ropes, respectively. (1)(2) are training rope before and after manipulation, (3)(4) are test rope before and after manipulation. Point registration runs on (c) between (1) and (3), obtaining a correspondence matrix $\mathbf{C}$, which is then used to multiply (d)(2), yielding (d)(4). (d)(4) is then converted back to Cartesian space, namely (b)(4), which is the test rope's target state.}
\label{fig:demo}
\end{figure}

Based on distances between point pairs, all the values in equation \eqref{eq:bayes} can be obtained. Define a correspondence matrix $\mathbf{C}_{m,n} = p(\mathbf{x}_{\theta n}|\mathbf{z}_{\theta m})$. The value of $\mathbf{C}_{m,n}$ reflects the degree to which $\mathbf{x}_{\theta n}$ is related to $\mathbf{z}_{\theta m}$, hence to $\mathbf{y}_{\theta m}$.

Next, determine the angle of each node on the target rope in test using a weighted average of the target rope during training. The desired tangent information for the test rope is 
\begin{align}
  \mathbf{X}_{\theta} = \mathbf{C}  \mathbf{Y}_{\theta}
  \label{eq:weighted_ave}
\end{align}

In order to determine where the gripper should go during test, the first step is to find the grasping node $\mathbf{y}_m$ on the training rope. $\mathbf{y}_m$ is the node closest to the gripper in training. Let $\mathbf{x}_n$ denote the grasping node for test, and it can be obtained through the following expression:
\begin{align}
  n = \operatorname*{arg\,max}_j \mathbf{C}_{m, j}
\end{align}
Further integration yields the Cartesian coordinates $(x_n, y_n)$ of node $\mathbf{x}_n$.
\begin{align}
  x_n = x_0 + \int_{0}^{n}dl\cdot cos\theta = x_0 + \sum_{i = 0}^{n} \delta l \cdot cos\mathbf{X}_{\theta_i}
\end{align}
\begin{align}
  y_n = y_0 + \int_{0}^{n}dl\cdot sin\theta = y_0 + \sum_{i = 0}^{n} \delta l \cdot sin\mathbf{X}_{\theta_i}
\end{align}
where $\delta l$ is the uniform length of each node on the rope, which is determined by the visual tracking system.

With this method, the length of the rope will always be conserved, since the gripper position is obtained by integration along the tangent direction of the rope.

Note that tangent space point set registration is not confined to coherent point drift method. As a general idea, any point set registration method can be applied. For example, the TPS-RPM\cite{TPS-RPM1} algorithm will directly give us a correspondence matrix needed in \eqref{eq:weighted_ave}, and we will be able to proceed from there. Different point set registration methods have their own advantages, here CPD is used as it can be generalized easily to any dimension and will enable us to do more complicated tasks in the future.

\section{Experiment Platform}
\subsection{Experiment Platform Architecture}
\begin{figure}[h]
\centering
\includegraphics[scale=0.6]{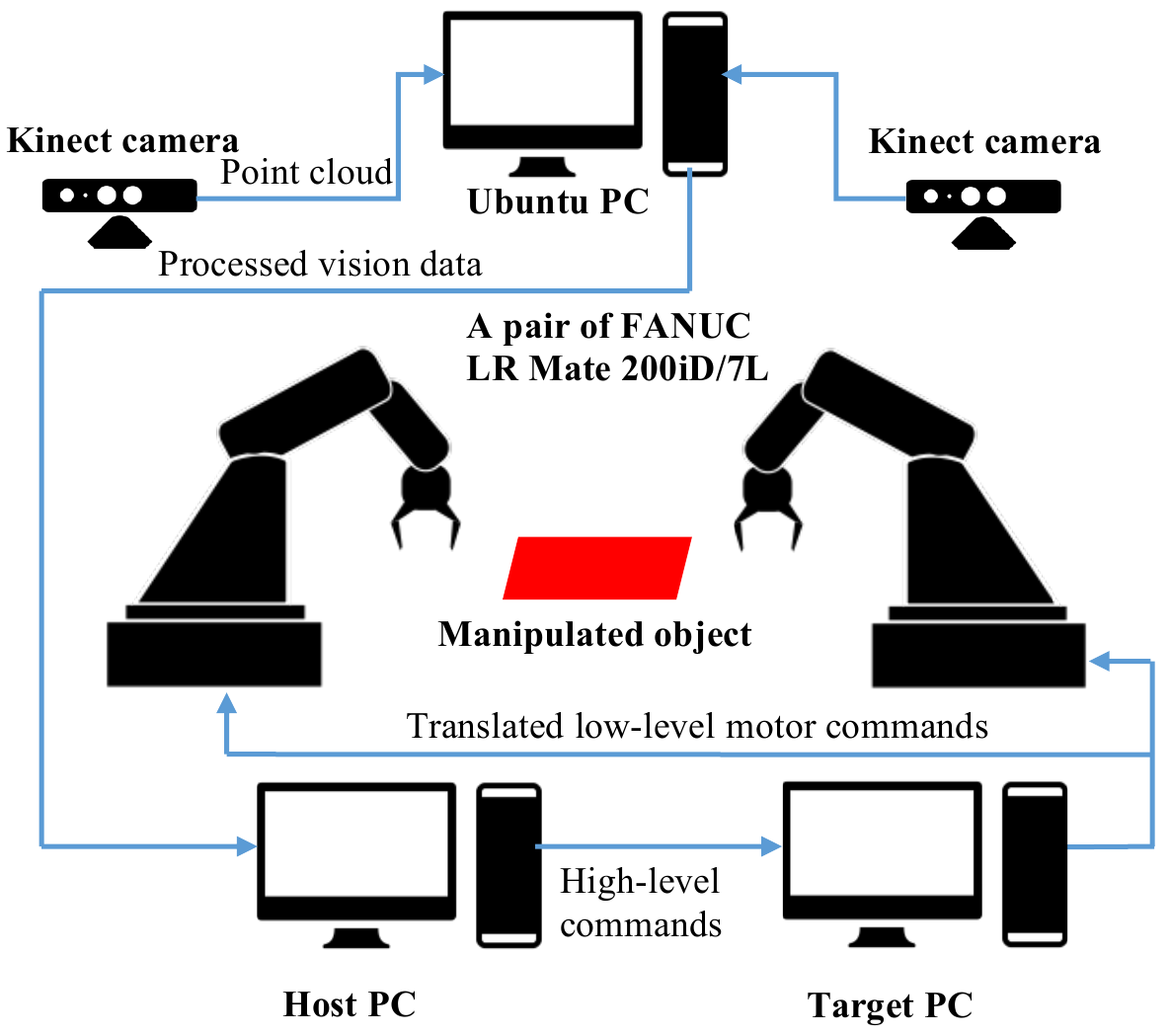}
\caption{
A schematic illustration of the experiment settings. The systems consists of cameras, three PCs, and a dual robot system.}
\label{fig:architecture}
\end{figure}

The experiment platform consists of a dual robot system and a vision system, as illustrated in Fig.~\ref{fig:architecture}. An experiment consists of training and testing.

In training, the robots enter a ``lead-through teaching'' mode where they can move according to forces applied to them. A human leads the robots through a trajectory, ``teaches'' them to operate, and records the information of several key frames, i.e. the position, attitude and status(open/closed) of two grippers. The vision system also records the state of the object before and after each operation. 

A schematic illustration of the experiment platform is shown in Fig.~\ref{fig:architecture}. Two FANUC LR Mate 200iD/7L robots are used to manipulate the object. Two Kinect Model 1517 cameras are placed in different directions to collect visual data, transmitting 640$\times$480 RGB-D image at 10Hz to Ubuntu PC, which runs a ROS\cite{ROS} node. Our vision algorithm\cite{tang2017state} abstracts the object as a set of nodes, and returns an array of Cartesian coordinates of all nodes at a slightly lower rate depending on the size of the object. The processed information, in the form of published ROS topic, is sent to Host PC, which runs the algorithm discussed above and generates the trajectory. Finally, the trajectory is sent to target PC, where the code that controls the robots is carried out.

\subsection{Parameters}
For coherent point drift algorithm, some parameters are tuned in order to achieve a balance between rigidity and precision. First, in order for the curves to be mapped together well, the scale of all points in $x$ and $y$ directions should be similar. In our experiment, the $x$ value is chosen to be the node index multiplied by $10$, while $y$ value is chosen to be the angle value in degrees.

The mapping function cannot be too arbitrary or it will directly map each point in training to testing evenly along the length, losing track of features such as corners. In experiment, the regularization coefficient $\lambda$ in  \eqref{eq:Qregularized} is chosen to be 10. Since outliers are already filtered out, the term for outliers, $\omega$, is set to be 0. The standard deviation of the Gaussian distribution for Green's function G in \eqref{eq:Green's function}, here denoted by $\beta$, is set to be 1. To save time without sacrificing the mapping precision, the maximum iteration for point set registration E-M step is set to be 100.

\section{Experiment Results}

Despite the fact that tangent space point set registration was first proposed to solve problems occurred in rope straightening, the algorithm is found to perform well on various tasks, such as cloth folding, cloth unfolding and knot tying. Two examples are shown in this section to offer a better understanding of its capability.

\subsection{Cloth Unfolding}
In this section, the performance of the proposed algorithm (point set registration in tangent space) with the previous method (point set registration in Cartesian space) will be compared for the task of cloth unfolding.

\begin{figure}[h]
\centering
\includegraphics[scale=0.55]{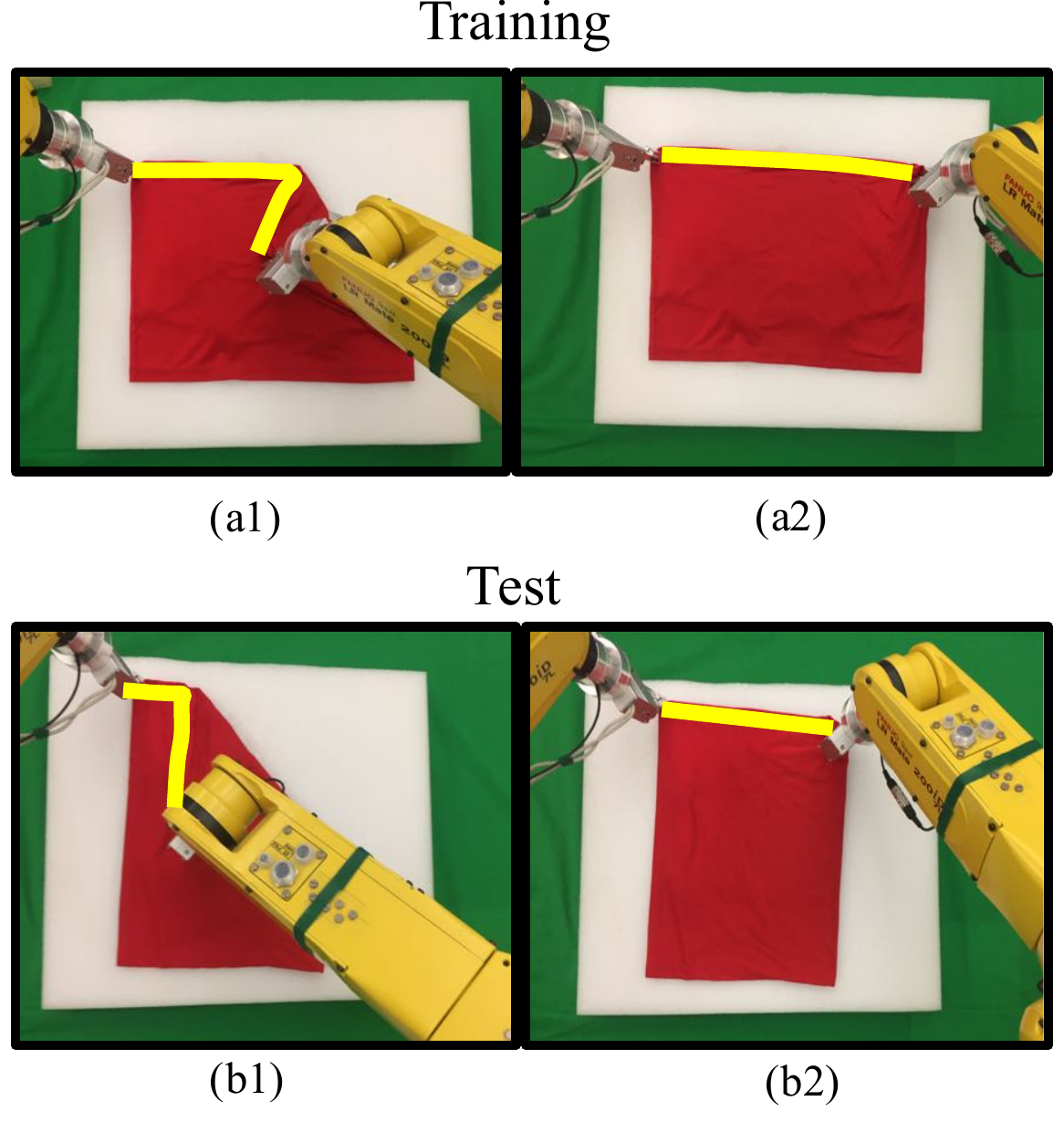}
\caption{
A cloth unfolding experiment. (a1), (a2) are the cloths in training, before and after manipulation. (b1), (b2) are the cloths in test, before and after manipulation. The upper edges, which we are most interested in, are marked in yellow.
}
\label{fig:cloth_exp} 
\end{figure}

For this experiment (Fig.~\ref{fig:cloth_exp}), in training, the cloth is 35cm in $y$ direction and 46cm in $x$ direction. In test, the cloth is rotated by 90 degrees, therefore it has a dimension of 46cm in $y$ and 35cm in $x$. Furthermore, the folding angles are also different.

In Fig.~\ref{fig:cloth_exp} (a1)(a2), the manipulation of unfolding the cloth is demonstrated to the robot by lead-through teaching. For cloth unfolding, we are interested in the upper edge of the cloth, which is marked in yellow. As is shown in Fig.~\ref{fig:cloth_exp}(b1)(b2), tangent space point set registration is able to unfold the cloth without overstretching or over-compressing. Note that the algorithm first extracts the upper edge of the cloth and runs on the upper edge.

\begin{figure}[h]
\centering
\includegraphics[scale=0.33]{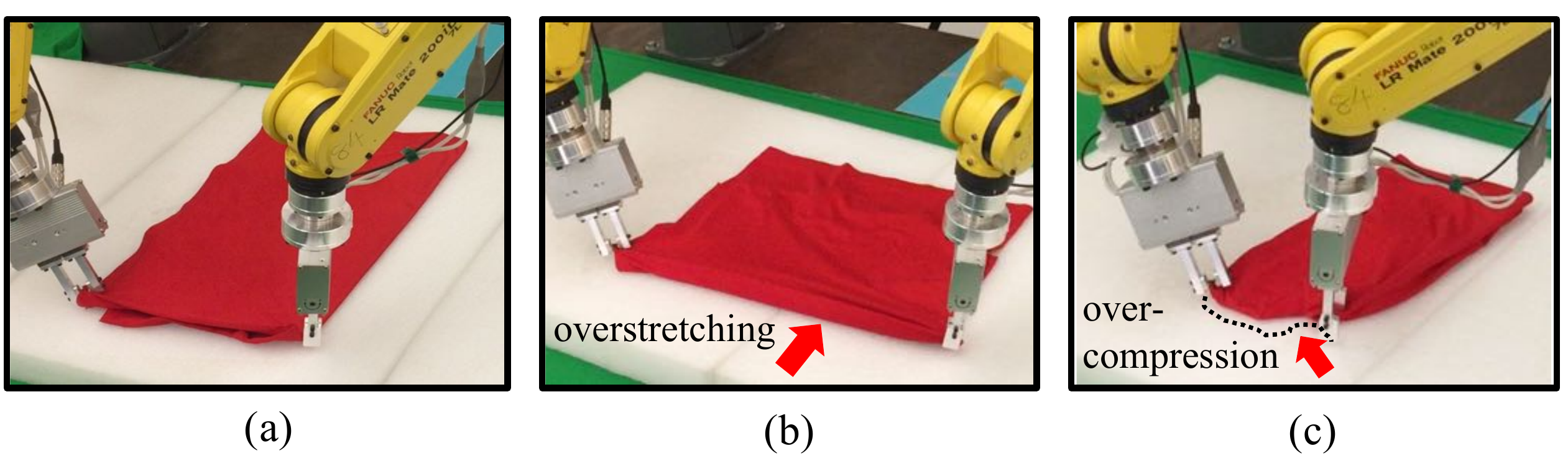}
\caption{ Common failures with Cartesian space point registration.
(a) The robot is trained to unfold the cloth. (b) Wider test cloth is overstretched. (c) Narrower test cloth is over-compressed.
}
\label{fig:cartesian_failure}
\end{figure}

On the other hand, Cartesian space point set registration fails to accomplish this task. In Fig.~\ref{fig:cartesian_failure}(c), the robots were faced with a narrow, folded cloth. In an effort to unfold it, they over-compress the cloth. Measurement shows that the upper edge shrinks in $x$ direction by 4cm, verifying our prediction in Fig.~\ref{fig_comparison1}. Furthermore, it has a 5cm shift in $y$ direction due to the warp of space. In experiment, these deviations resulted in a very wrinkled cloth, as shown in Fig.~\ref{fig:cartesian_failure}(c). Wider cloths, on the other hand, will be overstretched, as shown in Fig\ref{fig:cartesian_failure}(b).

\subsection{Rope Knotting}
Using point set registration in tangent space will yield a trajectory that can conserve the object's length, hence safer manipulation. So far, it has been shown that for cloth unfolding, point set registration in Cartesian space will not work. This also applies to any other manipulation that involves stretching an object to its full length.

In other scenarios, where formerly point mapping in Cartesian space succeeded, the proposed algorithm is also robust and safe. Rope knotting is the typical task the previous method was able to master (because the rope is unlikely to be fully stretched in a knotting task). Tangent space point mapping also works on rope knotting. A typical rope knotting task is divided into 4 steps, with 5 different rope states. A snapshot of each of these rope states is taken and tangent space point set registration is performed for every step.

\begin{figure}[h]
\centering
\includegraphics[scale=0.6]{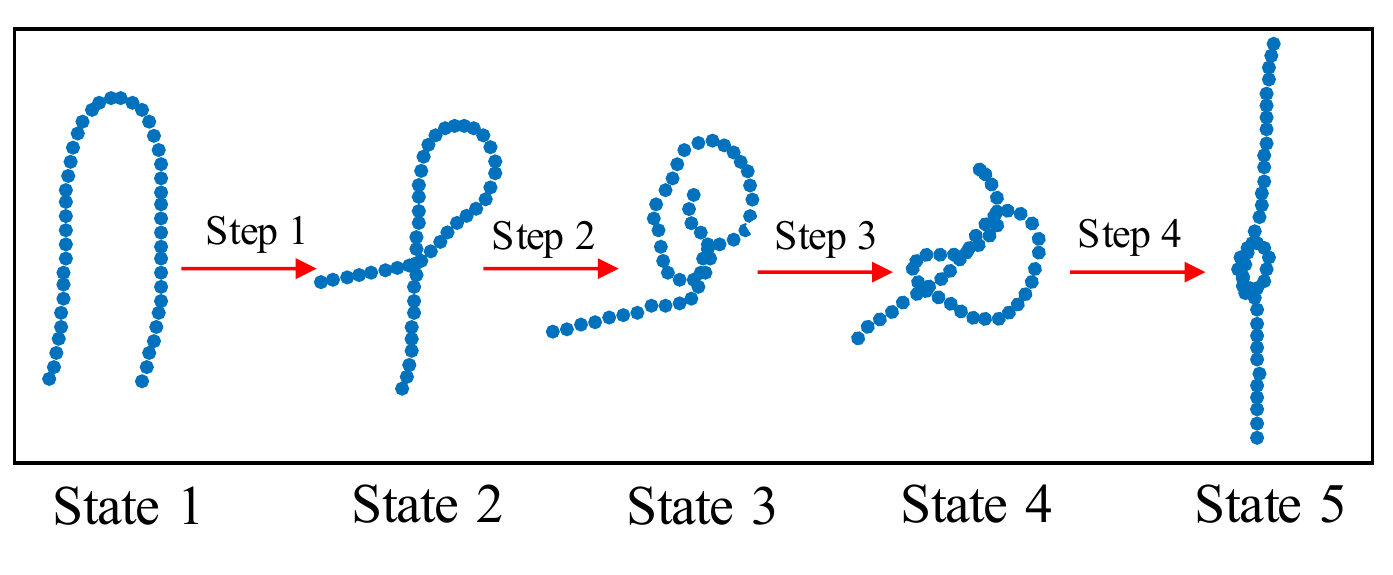}
\caption{
Illustration of all 5 states and 4 steps of a rope knotting task. All data is extracted from an experiment.
}
\label{fig:knot_steps}
\end{figure}

\begin{figure}[h]
\centering
\includegraphics[scale=0.25]{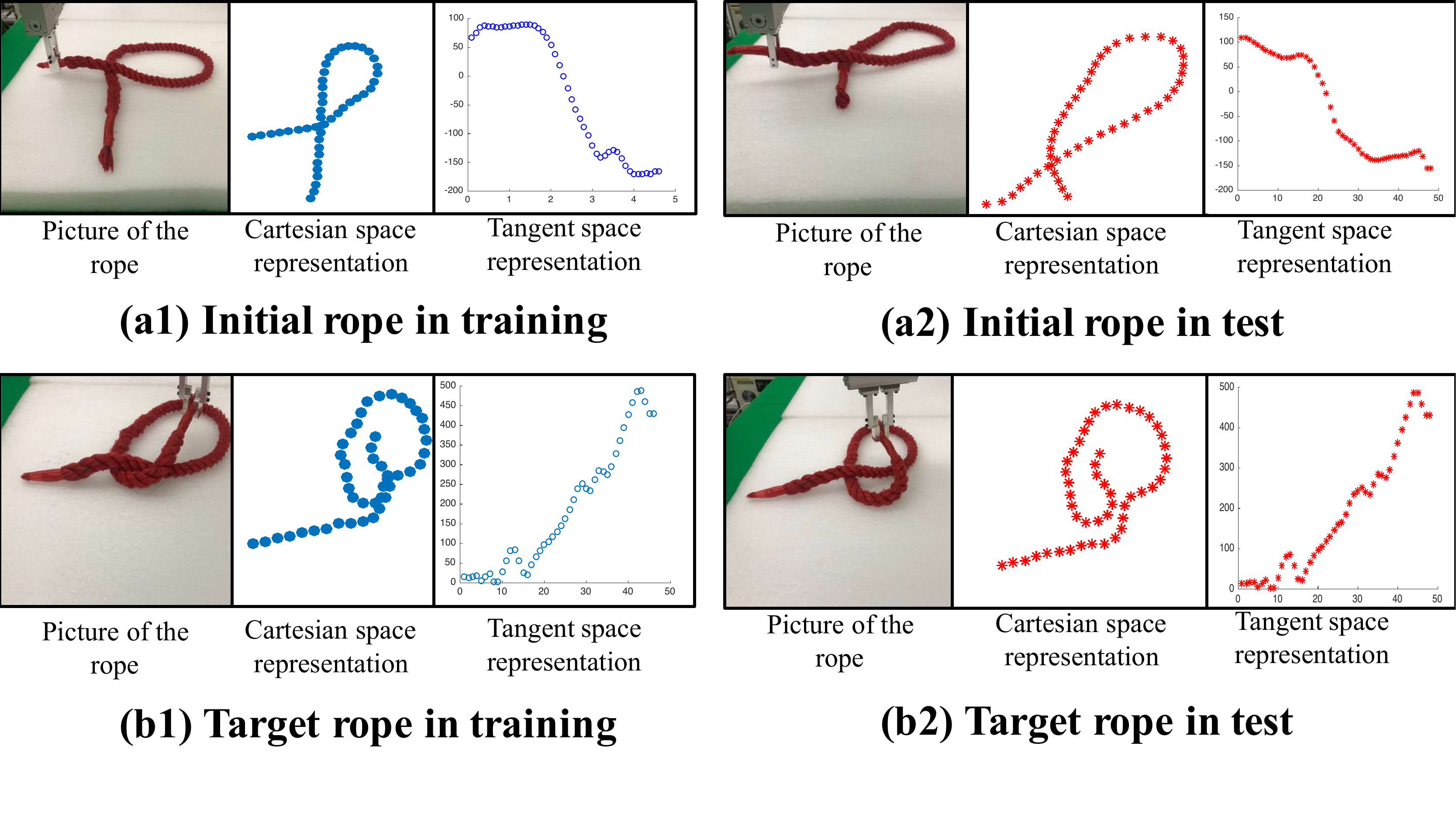}
\caption{
Step 2 of rope knotting.
}
\label{fig:knot_initial}
\end{figure}

As an example, Fig.~\ref{fig:knot_initial} demonstrates how the algorithm works in step 2.

Step by step, the robotic arms are able to carry out the entire task of knotting.
\section{Conclusion and Discussion}
In this work, point set registration in tangent space is proposed to transfer a robot's skill learned in training to test scenarios. Compared with the previous method which runs in Cartesian space, point set registration in tangent space is able to conserve the length of the manipulated object and satisfy physical constraints.

When it comes to 2-D object (such as cloth) manipulation, cases where the object is stretched becomes more common due to the increase in physical constraints. Therefore, the successful experiment results of tangent space point mapping indicate more possibilities for robots to operate on 2-D objects.

\addtolength{\textheight}{-12cm}   
\bibliographystyle{IEEEtran}
\bibliography{Reference}

\end{document}